%% file: main.tex
\documentclass{article} 
\usepackage{iclr2016_conference,times}
\usepackage{hyperref}
\usepackage{url}
\usepackage{graphicx}
\usepackage{xspace}
\usepackage{caption}
\usepackage{amsmath} 
\usepackage{amssymb}
\usepackage{framed}
\DeclareMathOperator*{\argmax}{arg\,max}

\usepackage{color}

\usepackage{algorithm}
\usepackage{algorithmic}

\newcommand{\N}{$N$}
\newcommand{\m}{{\bf m}}
\newcommand{\inp}{{x}} 

\DeclareMathOperator*{\memnn}{\text{MemNN}}
\DeclareMathOperator*{\ngram}{\text{N-gram}}
\DeclareMathOperator*{\svm}{\text{Structured SVM}}
\DeclareMathOperator*{\task}{\text{TASK}}
\definecolor{dgreen}{rgb}{0.0,0.4,0.0}
\definecolor{dred}{rgb}{0.7,0.0,0.0}

\newcommand{\G}[1]{\textcolor{dgreen}{#1 \tiny{ ex.}}}

\newcommand{\PT}{\textcolor{white}{100}}

\usepackage{mathtools}
\DeclarePairedDelimiter\ceil{\lceil}{\rceil}

\newcommand{\secsinglefact}{1}
\newcommand{\sectwofacts}{2}
\newcommand{\secthreefacts}{3}
\newcommand{\sectwoargrel}{4}
\newcommand{\secthreeargrel}{5}
\newcommand{\secyesno}{6}
\newcommand{\seccount}{7}
\newcommand{\secsets}{8}
\newcommand{\secnegation}{9}
\newcommand{\secindefreason}{10}
\newcommand{\seccoref}{11}
\newcommand{\secconjunction}{12}
\newcommand{\seccompoundcoref}{13}
\newcommand{\sectimereas}{14}
\newcommand{\secdeduction}{15}
\newcommand{\secinduction}{16}
\newcommand{\secposreas}{17}
\newcommand{\secszreas}{18}
\newcommand{\secpathfinder}{19}
\newcommand{\secagentsmotivations}{20}

\title{Towards AI-Complete Question Answering: \\A Set of Prerequisite Toy Tasks}

\author{Jason Weston, Antoine Bordes, Sumit Chopra, Alexander M. Rush, \\
 {\bf Bart van Merri\"enboer, Armand Joulin \& Tomas Mikolov}\\
Facebook AI Research\\
770 Broadway\\
New York, USA\\
\texttt{\{jase,abordes,spchopra,tmikolov,sashar,bartvm\}@fb.com}
}

\begin{document}

\maketitle

\begin{abstract}
One long-term goal of machine learning research is to produce methods that
are applicable to reasoning and natural language, in particular building
an intelligent dialogue agent.
To measure progress towards that goal, we argue for the usefulness of a set of proxy tasks
that evaluate reading comprehension via question answering.
Our tasks measure
understanding in several ways: whether a system is able to answer questions
via chaining facts, simple induction, deduction and many more.
The tasks  are designed to be prerequisites for any system that aims to be 
capable of conversing with a human.
We believe many existing learning systems
can currently not solve them, and hence our aim is to classify these tasks
into skill sets, so that researchers can identify (and then rectify) the failings of their systems.
We also extend and improve the recently introduced Memory Networks model, 
and show it is able to solve some, but not all, of the tasks.
\end{abstract}

\section{Introduction}

\input{Introduction.tex}

\section{Related Work}
\input{Related.tex}

\section{The Tasks} \label{sec:tasks}
\input{Tasks.tex}

\section{Simulation}\label{sec:simulation}
\input{Simulation.tex}

\section{Experiments} \label{sec:exp}

\input{Results.tex}

\section{Discussion} \label{sec:conclusion}

\input{Conclusion.tex}

\bibliography{refs}
\bibliographystyle{iclr2016_conference}

\appendix
\section{Extensions to Memory Networks} \label{sec:memnns}

\input{Models.tex}

\section{Baseline using External Resources} \label{sec:exp-nlp}

\input{NLPBaseline.tex}

\end{document}

%% file: Introduction.tex
There is a rich history of the use of synthetic tasks in machine learning, from the XOR problem which helped motivate neural networks \citep{minsky1969perceptron,rumelhart1985learning},
 to circle and ring datasets
that helped motivate some of the most well-known  clustering and semi-supervised learning
algorithms  \citep{ng2002spectral,zhu2003semi}, 
Mackey Glass equations for time series \citep{muller1997predicting},  and so on -- in fact some of the well known
UCI datasets \citep{uci} are synthetic as well (e.g., {\em waveform}).
Recent work continues this trend. For example, in the area of developing learning algorithms with a memory component
%
synthetic datasets were used to help develop both the Neural Turing Machine
of \citet{graves2014neural} and the Memory Networks of \citet{memnns},
%
the latter of which is relevant to this work.

One of the reasons for the interest in synthetic data is that it can be easier to develop new techniques using it.
It is well known that working with large amounts of real data (``big data'') 
tends to lead researchers to simpler models as ``simple models and a lot of data trump
more elaborate models based on less data'' \citep{halevy2009unreasonable}. 
For example, 
\N-grams for language modeling work well relative to existing competing methods, but 
are far from being a model that truly understands text.
As researchers we can become stuck in local minima in algorithm space;
development of synthetic data is one way to try and break out of that.

In this work we propose a framework and a set of synthetic tasks 
for the goal of helping to develop learning algorithms for text understanding and reasoning.
While it is relatively difficult to automatically evaluate the performance of an agent
in general dialogue -- a long term-goal of AI -- it is 
relatively easy to evaluate responses
 to input questions, i.e., the task of question answering (QA).
Question answering is incredibly broad:
 more or less any  task one can think of can be cast into this setup.
This enables us to propose  a wide ranging set of different tasks, that  test different capabilities of learning algorithms, under a common framework.

Our tasks are built with a unified underlying simulation of a physical world, akin to a classic text adventure game \citep{montfort2005twisty} whereby actors move around manipulating objects and interacting with each other. 
As the simulation runs, grounded text and question answer pairs are simultaneously generated. Our goal is to categorize {\em different  kinds of questions}
 into skill sets, which become our tasks. 
 Our hope is that the analysis of performance on these tasks will help expose weaknesses of current models and help motivate new algorithm designs that alleviate these weaknesses.
We further envision this as a feedback loop where  new tasks can then be designed in response, 
perhaps in an adversarial fashion, in order to break the new models. 

The tasks we design are detailed in Section \ref{sec:tasks}, and the simulation used to generate them in Section \ref{sec:simulation}. 
In Section \ref{sec:exp} we give 
benchmark results of standard methods on our tasks, and analyse their successes and failures.
In order to exemplify 
 the kind of
feedback loop between algorithm development and task development we envision,
in Section \ref{sec:memnns} we propose a set of improvements to the recent Memory Network method,
which has shown to give promising performance in QA. We show 
 our proposed approach
 does indeed give improved performance on some tasks, but is still unable to solve some of 
them, 
which we consider as open problems.




%% file: Related.tex
Several projects targeting language understanding using QA-based
strategies have recently emerged. 
Unlike tasks like dialogue or summarization, 
QA is easy to evaluate (especially in true/false or multiple choice scenarios) and
hence makes it an appealing research avenue.
The difficulty lies in the definition of questions: they must be
unambiguously answerable by adult humans (or children), but still
require some thinking.
The {\it Allen Institute for AI}'s flagship project {\sc
  aristo}\footnote{\tiny\url{http://allenai.org/aristo.html}} is organized
around a collection of QA tasks derived from
increasingly difficult science exams, at the 4th, 8th, and 12th grade
levels.
\citet{richardson2013mctest} proposed the \textcolor{black}{\it MCTest}\footnote{\tiny\url{http://research.microsoft.com/mct}}
a set of 660 stories and associated questions intended for research on
the machine comprehension of text. Each question requires the reader
to understand different aspects of the story.

These two initiatives go in a promising direction but interpreting the
results on these benchmarks remain complicated.
Indeed, no system has yet been able to fully solve the proposed tasks and
since many sub-tasks need to be solved to answer any of their questions
(coreference, deduction, use of common-sense, etc.), it is difficult to
clearly identify capabilities and limitations of these systems and
hence to propose improvements and modifications.
As a result, conclusions drawn from these projects are not much clearer
than that coming from more traditional works on QA over
large-scale Knowledge Bases 
\citep{berant2013semantic,fader2014open}.
Besides, the best performing systems are based on hand-crafted
patterns and features, and/or statistics acquired on very large
corpora. It is difficult to argue that such systems actually
understand language and are not simply light upgrades of traditional
information extraction methods \citep{yao2014freebase}.
The system of \citet{berant2014modeling} is more evolved since it
builds a structured representation of a text and of a question  
to answer. Despite its potential this method remains highly domain
specific and relies on a lot of prior knowledge.

Based on these observations, we chose to conceive a collection of much
simpler QA tasks, with the main objective that failure or success of a
system  on any of them can unequivocally \textcolor{black}{provide} feedback on its
capabilities.
In that, we are close to the {\it Winograd Schema Challenge}
\cite{levesque2011winograd}, which is organized around simple
statements followed by a single binary choice question such as: ``{\it Joan
  made sure to thank Susan for all the help she had received. Who had
  received the help?~Joan or Susan?}''.
In this challenge\textcolor{black}{, and our tasks,} it is straightforward to interpret
results.
Yet, where the Winograd Challenge is mostly centered around \textcolor{black}{evaluating} if
systems can acquire and make use of background knowledge that is not
expressed in the words of the statement, our tasks \textcolor{black}{are self-contained and
are more diverse. By self-contained we mean our tasks come  with
both training data and evaluation data, rather than just the latter as in the case of 
{\sc aristo} and the Winograd Challenge.
{\it MCTest} has a train/test split but the training set is likely too small to capture
all the reasoning needed to do well on the test set.
In our setup one can assess the amount of training examples needed to perform well
(which can be increased as desired)
and commonsense knowledge and reasoning required for the test set should be contained
in the training set.
In terms of diversity, some of our tasks are related to existing setups
but we also propose many additional ones;
tasks \secsets ~and \secnegation~ 
 are inspired by previous work on lambda dependency-based compositional semantics
\citep{liang2013learning,liang2013lambda} for instance.
For us, each task checks one skill that the system must have and we
postulate that performing well on all of them is a prerequisite 
for any system aiming at full text understanding and reasoning.
}
%

%% file: Tasks.tex
\paragraph{Principles}
Our main idea is to provide a set of tasks, in a similar way to
how software testing is built in computer science. Ideally each task is a
``leaf'' test case, as independent from others as possible, 
and tests in the simplest way possible one aspect of intended behavior. Subsequent (``non-leaf'') tests can build on these by testing 
 combinations as well.
The tasks are publicly available at ~\url{http://fb.ai/babi}.
Source code to generate the tasks is available at
\url{https://github.com/facebook/bAbI-tasks}.

Each task provides a set of training and test data, with the
intention that a successful model performs well on test data. 
Following \cite{memnns}, the supervision in the training set is given by the true answers to questions, and the set of {\em relevant} statements for answering a given question, which may or may not be used by the learner.
We set up the tasks so that correct answers are limited to a single word ({\em Q: Where is Mark? A: bathroom}), 
or else a list of words ({\em Q: What is Mark holding?}) as evaluation is then clear-cut, and is measured simply as right or wrong.

All of the tasks are noiseless and a human able to read that language
can 
potentially 
achieve 100\% accuracy. 
We tried to 
choose tasks that are natural to a human: they are
based on simple usual situations and
 no background in 
areas such as formal semantics, machine learning, logic or
knowledge representation is required for an adult to solve them. 

The data itself is produced using a simple simulation of characters and objects moving around and interacting in locations, described in Section
\ref{sec:simulation}. The simulation allows us to generate data in many different scenarios  where the true labels are known by grounding to the simulation.
For each task, we describe it by giving a small sample of the dataset including statements, questions and the true labels (in red) in 
Tables \ref{taskseta} and \ref{tasksetb}.

\begin{table*}[t]
\begin{small}
\caption{Sample statements and questions from tasks 1 to 10.  \label{taskseta}}\vspace{2mm}
\begin{tabular}{|l|c|l|}
\cline{1-1}\cline{3-3}
&& \\[-2ex]
{\bf Task 1: Single Supporting Fact}         &&   {\bf Task 2: Two Supporting Facts} \\
&& \\[-2ex]
~~Mary went to the bathroom.                   &&~~John is in the playground.\\
~~John moved to the hallway.                    &&~~John picked up the football.\\
~~Mary travelled to the office.               &&~~Bob went to the kitchen.\\
~~Where is Mary?  \textcolor{red}{A:office}   &&~~Where is the football?  \textcolor{red}{A:playground} \\
\cline{1-1}\cline{3-3} 
\multicolumn{3}{c}{}
\\ 
\cline{1-1}\cline{3-3}
&& \\[-2ex]
{\bf Task 3: Three Supporting Facts}          &&   {\bf Task 4: Two Argument Relations} \\
&& \\[-2ex]
~~John picked up the apple.                     &&~~The office is north of the bedroom.\\
~~John went to the office. &&~~The bedroom is north of the bathroom.\\
~~John went to the kitchen. &&~~The kitchen is west of the garden.\\
~~John dropped the apple.  &&~~What is north of the bedroom? \textcolor{red}{A: office}\\
~~Where was the apple before the kitchen? \textcolor{red}{A:office} &&~~What is the bedroom north of? \textcolor{red}{A: bathroom} \\
\cline{1-1}\cline{3-3}
\multicolumn{3}{c}{}
\\ 
\cline{1-1}\cline{3-3}
&& \\[-2ex]
{\bf Task 5: Three Argument Relations}          &&   {\bf Task 6: Yes/No Questions} \\
&& \\[-2ex]
~~Mary gave the cake to Fred.  && ~~John moved to the playground.\\
~~Fred gave the cake to Bill.  && ~~Daniel went to the bathroom. \\
~~Jeff was given the milk by Bill. && ~~John went back to the hallway. \\
~~Who gave the cake to Fred? \textcolor{red}{A: Mary} && ~~Is John in the playground? \textcolor{red}{A:no} \\
~~Who did Fred give the cake to? \textcolor{red}{A: Bill} && ~~Is Daniel in the bathroom? \textcolor{red}{A:yes}\\ 
\cline{1-1}\cline{3-3}
\multicolumn{3}{c}{}
\\ 
\cline{1-1}\cline{3-3}
&& \\[-2ex]
{\bf Task 7:  Counting}          &&   {\bf Task 8: Lists/Sets} \\
&& \\[-2ex]
~~Daniel picked up the football. && ~~Daniel picks up the football.\\
~~Daniel dropped the football. && ~~Daniel drops the newspaper.\\
~~Daniel got the milk. && ~~Daniel picks up the milk.\\
~~Daniel took the apple. && ~~John took the apple. \\
~~How many objects is Daniel holding? \textcolor{red}{A: two} && ~~What is Daniel holding? \textcolor{red}{milk, football} \\
\cline{1-1}\cline{3-3}
\multicolumn{3}{c}{}
\\ 
\cline{1-1}\cline{3-3}
&& \\[-2ex]
{\bf Task 9: Simple Negation}          &&   {\bf Task 10: Indefinite Knowledge} \\
&& \\[-2ex]
~~Sandra travelled to the office. && ~~John is either in the classroom or the playground. \\
~~Fred is no longer in the office. && ~~Sandra is in the garden.\\
~~Is Fred in the office?   \textcolor{red}{A:no} && ~~Is John in the classroom?  \textcolor{red}{A:maybe} \\
~~Is Sandra in the office?   \textcolor{red}{A:yes} && ~~Is John in the office?  \textcolor{red}{A:no} \\
\cline{1-1}\cline{3-3}
\end{tabular}
\end{small}
\vspace*{-3ex}
\end{table*}

\begin{table*}[t]
\caption{Sample statements and questions from tasks 11 to 20.  \label{tasksetb}}\vspace{2mm}
\begin{small}
\begin{tabular}{|l|c|l|}
\cline{1-1}\cline{3-3}
&& \\[-2ex]
{\bf Task 11: Basic Coreference}          &&   {\bf Task 12: Conjunction} \\
&& \\[-2ex]
~~Daniel was in the kitchen. && ~~Mary and Jeff went to the kitchen.\\
~~Then he went to the studio. && ~~Then Jeff went to the park.\\
~~Sandra was in the office. && ~~Where is Mary? \textcolor{red}{A: kitchen} \\
~~Where is Daniel?  \textcolor{red}{A:studio} && ~~Where is Jeff? \textcolor{red}{A: park} \\
\cline{1-1}\cline{3-3}
\multicolumn{3}{c}{}
\\ 
\cline{1-1}\cline{3-3}
&& \\[-2ex]
{\bf Task 13: Compound Coreference}          &&   {\bf Task 14: Time Reasoning} \\
&& \\[-2ex]
~~Daniel and Sandra journeyed to the office. && ~~ In the afternoon Julie went to the park. \\
~~Then they went to the garden. && ~~Yesterday Julie was at school.\\
~~Sandra and John travelled to the kitchen. && ~~Julie went to the cinema this evening.\\
~~After that they moved to the hallway. && ~~Where did Julie go after the park? \textcolor{red}{A:cinema} \\
~~Where is Daniel? \textcolor{red}{A: garden} && ~~Where was Julie before the park? \textcolor{red}{A:school} \\
\cline{1-1}\cline{3-3}
\multicolumn{3}{c}{}
\\ 
\cline{1-1}\cline{3-3}
&& \\[-2ex]
{\bf Task 15: Basic Deduction}          &&   {\bf Task 16: Basic Induction} \\
&& \\[-2ex]
~~Sheep are afraid of wolves. && ~~Lily is a swan. \\
~~Cats are afraid of dogs. && ~~Lily is white. \\
~~Mice are afraid of cats. && ~~Bernhard is green.\\
~~Gertrude is a sheep. && ~~Greg is a swan.\\
~~What is Gertrude afraid of? \textcolor{red}{A:wolves} && ~~What color is Greg? \textcolor{red}{A:white} \\ 
\cline{1-1}\cline{3-3}
\multicolumn{3}{c}{}
\\ 
\cline{1-1}\cline{3-3}
&& \\[-2ex]
{\bf Task 17: Positional Reasoning}       &&   {\bf Task 18: Size Reasoning} \\
&& \\[-2ex]
~~The triangle is to the right of the blue square.&& ~~The football fits in the suitcase.\\
~~The red square is on top of the blue square.&& ~~The suitcase fits in the cupboard.\\
~~The red sphere is to the right of the blue square. && ~~The  box is smaller than the football.\\
~~Is the red sphere to the right of the blue square? \textcolor{red}{A:yes} && ~~Will the box fit in the suitcase? \textcolor{red}{A:yes} \\
~~Is the red square to the left of the triangle? \textcolor{red}{A:yes} && ~~Will the cupboard fit in the box?  \textcolor{red}{A:no} \\
\cline{1-1}\cline{3-3}
\multicolumn{3}{c}{}
\\ 
\cline{1-1}\cline{3-3}
&& \\[-2ex]
{\bf Task 19: Path Finding}          &&   {\bf Task 20: Agent's Motivations} \\
&& \\[-2ex]
~~The kitchen is north of the hallway. && ~~{John is hungry.}\\
~~The bathroom is west of the bedroom.  && ~~{John goes to the kitchen.}\\
~~The den is east of the hallway. && ~~John grabbed the apple there.\\
~~The office is south of the bedroom.  && ~~{Daniel is hungry.}\\
~~How do you go from den to kitchen? \textcolor{red}{A: west, north} && ~~{Where does Daniel go?}  \textcolor{red}{A:kitchen}\\
~~How do you go from office to bathroom? \textcolor{red}{A: north, west} && ~~{Why did John go to the kitchen?}  \textcolor{red}{A:hungry}\\
\cline{1-1}\cline{3-3}
\end{tabular}
\end{small}
\vspace*{-3ex}
\end{table*}

\vspace{-2mm}
\paragraph{Single Supporting Fact}
Task \secsinglefact~
consists of questions where a previously given single supporting fact, 
 potentially amongst a set of other irrelevant facts, provides the answer. We first test one of the simplest cases of this, by asking for the location of a person, e.g. {\em ``Mary travelled to the office. Where is Mary?''}. 
This kind of task was already employed in \citet{memnns}. 
It can be considered the simplest case of some real world QA datasets such as in \citet{paralex}.

\vspace{-2mm}
\paragraph{Two or Three Supporting Facts}
A harder task is to answer questions where two supporting statements have to be chained to answer the question, as in task \sectwofacts, where to answer the question {\em ``Where is the football?''} one has to combine information from two sentences {\em ``John is in the playground''} and {\em ``John picked up the football''}. Again, this kind of task was already used in \cite{memnns}. 
Similarly, one can make a task with three supporting facts, given in task \secthreefacts,
whereby  the first three statements are all required to answer the question
``Where was the apple before the kitchen?''.

\vspace{-2mm}
\paragraph{Two or Three Argument Relations}
To answer questions the ability to differentiate and recognize subjects and objects is crucial. 
In task \sectwoargrel~we consider  the extreme case where sentences feature re-ordered words,
i.e. a bag-of-words will not work. For example, the questions
{\em ``What is north of the bedroom?''} and {\em ``What is the bedroom north of?''} have 
exactly the same words, but a different order, with different answers.
A step further, sometimes one needs to  differentiate three 
separate arguments. Task 5 involves statements like 
{\em ``Jeff was given the milk by Bill''}
 and then queries who is the giver, receiver or which object is involved. 

\vspace{-2mm}
\paragraph{Yes/No Questions}
Task \secyesno~tests, on some of the simplest questions possible (specifically, ones with a single supporting fact), the ability of a model to answer true/false type questions like {\em ``Is John in the playground?''}.

\vspace{-2mm}
\paragraph{Counting and Lists/Sets}
Task \seccount~tests the ability of the QA system to perform simple counting operations, by asking about the number of objects with a certain property, e.g.
 {\em ``How many objects is Daniel holding?''}.
Similarly, task \secsets~tests the ability to produce a set of single word answers in the form of a list, e.g. {\em ``What is Daniel holding?''}.
These tasks can be seen as  QA tasks related to basic database search operations.

\vspace{-2mm}
\paragraph{Simple Negation and Indefinite Knowledge}
Tasks 9 and 10 test 
slightly more complex natural language constructs.
Task \secnegation~tests one of the simplest forms of negation, that of supporting facts that imply a statement is false e.g.  {\em ``Fred is no longer in the office''} rather than {\em ``Fred travelled to the office''}.
(In this case, task \secyesno~(yes/no questions) is  a prerequisite to the task.)
Task \secindefreason~tests if we can  model statements that describe possibilities rather than certainties, e.g. {\em ``John is either in the classroom or the playground.''},  where in that case the answer is {\em ``maybe''} to the question {\em ``Is John in the classroom?''}.

\vspace{-2mm}
\paragraph{Basic Coreference, Conjunctions and Compound Coreference}
Task \seccoref~tests the simplest type of coreference, that of detecting the nearest referent, e.g. {\em ``Daniel was in the kitchen. Then he went to the studio.''}. 
Real-world data typically addresses this as a labeling problem
and studies more sophisticated phenomena \citep{soon2001machine}, whereas we evaluate it as
in all our other tasks as a question answering problem.
Task \secconjunction~(conjunctions) tests referring to multiple subjects in a single
 statement, e.g. {\em ``Mary and Jeff went to the kitchen.''}.
Task \seccompoundcoref~tests coreference in the case where the pronoun can refer to multiple actors, e.g. {\em ``Daniel and Sandra journeyed to the office. Then they went to the garden''}.

\vspace{-2mm}
\paragraph{Time Reasoning}
While our tasks so far have included time implicitly in the {\em order} of the statements,  task 
\sectimereas~tests understanding the use of time expressions within the statements,
e.g. {\em ``In the afternoon Julie went to the park. Yesterday Julie was at school.''}, followed by questions about the order of events such as {\em ``Where was Julie before the park?''}.
Real-world datasets address the task
of evaluating time expressions  typically as a labeling, rather than a QA task,
see e.g.  \cite{uzzaman2012tempeval}.

\vspace{-2mm}
\paragraph{Basic Deduction and Induction}
Task \secdeduction~tests basic deduction via inheritance of properties, e.g.
{\em ``Sheep are afraid of wolves. Gertrude is a sheep. What is Gertrude afraid of?''}.
Task \secinduction~similarly tests basic induction via 
inheritance of properties.
A full analysis of induction and deduction is clearly beyond the scope of this work,
and future tasks should analyse further, deeper aspects.

\vspace{-2mm}
\paragraph{Positional and Size Reasoning}
Task \secposreas~tests spatial reasoning, one of many components of the classical SHRDLU system \citep{winograd1972understanding} by asking questions about the relative positions of colored blocks. Task \secszreas~requires reasoning about the relative size of objects and is inspired
by the commonsense reasoning examples in the Winograd schema challenge 
\citep{levesque2011winograd}.

\vspace{-2mm}
\paragraph{Path Finding}
The goal of task \secpathfinder~is to find the path between locations: given the description of various locations, it asks: how do you get from one to another? This is related to the work of \cite{chen2011learning} and effectively involves a search problem.

\vspace{-2mm}
\paragraph{Agent's Motivations}
Finally, task \secagentsmotivations~questions, in the simplest way possible,
{\em why} an agent performs an action. 
It addresses the case of actors being in a given state (hungry, thirsty, tired, \dots) and the actions they then take, e.g. it should learn that hungry people might go to the kitchen, and so on. 


As already stated, these tasks are meant to foster the development and understanding of machine
learning algorithms.
A single model should be evaluated across all the tasks (not tuning per task) and then the same 
model should be tested on additional real-world tasks.

In our data release, in addition to providing the above 20 tasks in English, we also provide them
(i) in Hindi; and (ii) with shuffled English words so they are no longer readable by humans.
A good learning algorithm should  perform similarly on all three, which would
likely not be the case for a method using external resources, 
a setting intended to
mimic a learner being first presented a language and having to learn from scratch.


%% file: Simulation.tex

All our tasks are generated with a  simulation which behaves like a classic text adventure game.
The idea is that generating text within this simulation allows us to
ground the language used into a coherent and controlled (artificial) world.
Our simulation follows those of \cite{bordes2010towards,memnns} but
is somewhat more complex. 

The simulated world is composed of entities of various types
(locations, objects, persons. etc.) and of various
actions that operate on these entities. 
Entities have internal states: their location, whether they carry
objects on top or inside them (e.g., tables and boxes), the mental state of actors (e.g. hungry),
as well as properties such as size, color, and edibility.  
For locations, the nearby places that are connected (e.g. what lies to the east, or above) are encoded.
For actors, a set of pre-specified rules per actor can also be specified to 
control their behavior, e.g. if they are hungry they may try to find food. Random valid actions can also be executed if no rule is set, e.g. walking around randomly.

The actions an actor can execute in the simulation consist of the following:
{\it go $<$location$>$}, {\em get $<$object$>$}, {\em get $<$object1$>$ from $<$object2$>$}, {\em put $<$object1$>$ in/on $<$object2$>$}, {\em give $<$object$>$ to $<$actor$>$}, {\em drop $<$object$>$}, {\em set $<$entitity$>$ $<$state$>$}, {\em look}, {\em inventory} and {\em examine $<$object$>$}.
%
%
A set of universal constraints is imposed on those actions to enforce
coherence in the simulation. For example an actor cannot get something that
they or someone else already has, they cannot go to a place that is not connected to the current location, cannot drop
something they do not already have, and so on.  
%
Using the underlying actions, rules for actors, and their constraints, defines how
actors act. 
For each task we limit the actions needed for that task, e.g. task 1 only needs {\em go} whereas
task 2 uses {\em go}, {\em get} and {\em drop}.
%
If we write the commands down 
this gives us a very simple ``story''
which is executable by the simulation, e.g., 
{\em { joe go playground}; {bob  go office}; {joe get football}}. 
This example corresponds to task 2.
The system can then ask questions about the state of the simulation e.g.,
{\em where john?}, {\em where football?} and so on.
It is easy to calculate the true answers for these questions as we have access to
the underlying world.

To produce more natural looking text with lexical variety from statements and
questions we employ a simple
automated grammar. Each verb is assigned a set of synonyms,
e.g., the simulation command {\em get} is replaced with  either
{\em picked up}, {\em got}, {\em grabbed} or {\em took}, and {\em drop} is replaced with 
either {\em dropped}, {\em left}, {\em discarded}  or {\em put down}.
Similarly, each object and actor can have a set of replacement synonyms as well, e.g. replacing {Daniel} with {\em he} in task 11.
%
Adverbs are crucial for some tasks such as the time reasoning
task~14.
%

There are a great many aspects of language not yet modeled. For
example, all sentences are so far relatively short and contain
little nesting. Further, the entities and the vocabulary size is small  (150 words, and typically 4 actors, 6 locations and 3 objects used per task).
%
The hope is that defining a set of well defined tasks will help evaluate models in a controlled way within the simulated environment, which is hard to do with real data. 
That is, these tasks are not a substitute for real data, but should complement
them, especially when developing and analysing algorithms.
%

%% file: Results.tex
 \begin{table*}[t]
\caption{\small Test accuracy (\%) on our 20 Tasks for various methods 
(1000 training examples each). 
Our proposed extensions to MemNNs are in columns 5-9: with adaptive memory (AM), \N-grams (NG),
nonlinear matching function (NL), and combinations thereof.
Bold numbers indicate tasks where our extensions achieve $\ge 95\%$ accuracy
 but the original MemNN model of \citet{memnns} did not.
The last two columns (10-11) give extra analysis of the
$\memnn\limits_{\textcolor{black}{\tiny\mbox{\em AM + NG + NL}}}$ method.
Column 10 gives the amount of training data for each task needed
 to obtain $\geq 95\%$ accuracy, or {\em FAIL} if this is not achievable with 1000 training examples. 
The final column gives the accuracy when training 
on all data at once, rather than separately.
\label{table:results}}
\resizebox{1.01\linewidth}{!}{
 \begin{tabular}{l|c@{\,~}|@{\,~}c@{\,~}||@{\,}c@{\,}||@{\,}c@{\,}|@{\,}c@{\,}|@{\,}c@{\,}|@{\,}c@{\,}|@{\,}c@{\,}@{\,}r@{\,~~}@{\,}c@{\,~}|}
\cline{2-11}
 &\multicolumn{2}{c|}{{Weakly}} 
 &\multicolumn{1}{@{\,}c@{\,}|}{Uses External} 
 &\multicolumn{7}{@{\,}c@{\,}|}{Strong Supervision} \\
 &\multicolumn{2}{@{\,}c@{\,}|}{{Supervised}} 
 &\multicolumn{1}{@{\,}c@{\,}|}{Resources} 
 &\multicolumn{7}{@{\,}c@{\,}|}{(using supporting facts)} \\
\cline{2-11}
     {\rotatebox[origin=l]{0}{~~~~~~~~~~~~~~~~~~~~$\task\limits_{\mbox{~~~~~~~~~}}$}}
   &  {\rotatebox[origin=l]{60}{$\ngram\limits_{\textcolor{black}{\tiny\mbox{Classifier}}}$}} &
     {\rotatebox[origin=l]{60}{LSTM}} &  
     {\rotatebox[origin=l]{60}{$\svm\limits_{\textcolor{blue}{\tiny\mbox{\em COREF+SRL features}}}$}} &
     {\rotatebox[origin=l]{60}{$\memnn\limits_{{\tiny\mbox{\cite{memnns}}}} $}} &  
     {\rotatebox[origin=l]{60}{$\memnn\limits_{\textcolor{blue}{\tiny\mbox{\em ADAPTIVE MEMORY}}}$}} &  
     {\rotatebox[origin=l]{60}{$\memnn\limits_{\textcolor{blue}{\tiny\mbox{\em AM + N-GRAMS}}}$}} &
     {\rotatebox[origin=l]{60}{$\memnn\limits_{\textcolor{blue}{\tiny\mbox{\em AM + NONLINEAR}}}$}} &
     {\rotatebox[origin=l]{60}{$\memnn\limits_{\textcolor{blue}{\tiny\mbox{\em AM + NG + NL}}}$}} &
     {\rotatebox[origin=l]{60}{\textcolor{dgreen}{\tiny{\bf No. of ex. req. $\geq$ 95}}}} &
     {\rotatebox[origin=l]{60}{\textcolor{dgreen}{\tiny{MultiTask Training}}}} \\
1 - Single Supporting Fact &  36 & 50 & 99  & 100 & 100 & 100 & 100 & 100 & \G{250} & 100 \\  
2 - Two Supporting Facts &  2 & 20 & 74 & 100 & 100 & 100 & 100 & 100 & \G{500} & 100 \\  
3 - Three Supporting Facts & 7 & 20 & 17  & 20 & {\bf 100} &{\bf 99} & {\bf 100} &  {\bf 100}  & \G{500} & {\bf 98} \\ 
4 - Two Arg. Relations &  50 & 61 & 98 & 71 & 69 & {\bf 100} & 73 & {\bf 100}    &  \G{500} & 80 \\  
5 - Three  Arg. Relations &  20  & 70  & 83 & 83 & 83 & 86  & 86 & {\bf 98}   & \G{1000} & {\bf 99} \\  
6 - Yes/No Questions &  49 & 48 & 99 & 47 & 52 & 53 & {\bf 100} &  {\bf 100}    & \G{500} & {\bf 100} \\ 
7 - Counting &  52  & 49 & 69 & 68 & 78  & 86 & 83 &  85  &  \textcolor{red}{FAIL~~}  & 86 \\    
8 - Lists/Sets &  40 & 45 & 70  & 77 & 90 & 88 & 94 &  91 &  \textcolor{red}{FAIL~~}  & 93 \\ 
9 - Simple Negation & 62 & 64 & 100  & 65 & 71 & 63 & {\bf 100}  & {\bf 100 } & \G{500} & {\bf 100} \\  
10 - Indefinite Knowledge & 45 & 44 & 99 & 59 & 57 & 54 & {\bf 97}  &  {\bf 98}   & \G{1000} & {\bf 98} \\  
11 -  Basic Coreference           & 29 & 72 & 100 & 100 & 100 & 100 & 100 & 100 & \G{250} & 100 \\  
12 - Conjunction             & 9 & 74 & 96 & 100 & 100 & 100 & 100 & 100 & \G{250} & 100 \\  
13 - Compound Coref.  & 26 & 94 & 99  & 100 & 100 & 100 & 100  & 100 & \G{250} & 100 \\ 
14 - Time Reasoning & 19  & 27 & 99 & 99 & 100 & 99 & 100 &  99  & \G{500} & 99 \\  
15 - Basic Deduction & 20 & 21 & 96 & 74 & 73 & {\bf 100} & 77 &  {\bf 100} & \G{100} & {\bf 100} \\  
16 - Basic Induction & 43 & 23 & 24  & 27 & {\bf 100} & {\bf 100}  & {\bf 100} & {\bf 100} & \G{100} & 94 \\  
17 - Positional Reasoning & 46 & 51 & 61  & 54 & 46 & 49 & 57 & 65 & \textcolor{red}{FAIL~~} & 72 \\  
18 - Size Reasoning & 52 & 52 & 62 & 57 & 50 & 74 & 54 & {\bf 95} & \G{1000} & 93 \\  
19 - Path Finding & 0 & 8 & 49 & 0 & 9 & 3 & 15 &  36  & \textcolor{red}{FAIL~~} & 19 \\  
20 - Agent's Motivations & 76 & 91 & 95 & 100 & 100 &  100 & 100 & 100  & \G{250} & 100 \\ 
%
%
%
\hline
Mean Performance  & 34 & 49 & 79  & 75 & 79  & 83 & 87 & 93 & \PT & 92 \\ 
\cline{1-11}
\end{tabular}
}
\end{table*}

We compared the following methods on our tasks (on the English dataset):
(i) an \N-gram classifier baseline, (ii) LSTMs (long short term memory Recurrent Neural Networks) \mbox{\citep{hochreiter1997long}}, (iii) Memory Networks (MemNNs) \citep{memnns},
(iv) some extensions of Memory Networks we will detail; and (v) a structured SVM
that incorporates external labeled data from existing NLP tasks.
These models belong to three separate tracks.
Weakly supervised models are only given question answer pairs at training time, whereas strong supervision provides the set of supporting facts at training time (but not testing time) as well.
Strongly supervised ones  give accuracy upper bounds for weakly supervised models, i.e. the performance should be superior given the same model class.
Methods in the last external resources track can use labeled data from other sources rather than just the training set provided, e.g. coreference and semantic role labeling tasks, as well as strong supervision.
%
For each task we use 1000 questions for training, and 1000 for testing, and report the test accuracy.
We consider a task successfully passed if $\geq 95\%$ accuracy is obtained\footnote{The choice of 95\% (and 1000 training examples) is arbitrary.}.

\paragraph{Methods}

The \N-gram classifier baseline is inspired by the baselines in \mbox{\citet{richardson2013mctest}} but applied to the case of producing a 1-word answer rather than a multiple choice question: we construct a bag-of-\N-grams for all sentences in the story that share at least one word with the question, and then learn a linear classifier to predict the answer using those features\footnote{Constructing \N-grams from all sentences rather than using the filtered set gave worse results.}.

LSTMs are a popular method for sequence prediction \citep{sutskever2014sequence} and outperform standard RNNs (Recurrent Neural Networks) for tasks similar to ours \citep{memnns}. They work by reading the story 
until the point they reach a question and then have to output an answer.
Note that they are weakly supervised by answers only, 
and are hence at a disadvantage compared to strongly supervised methods or methods
that use external resources.

MemNNs \citep{memnns} are a recently proposed class of models that have been shown to perform well at QA.
They work by a ``controller'' neural network performing inference over the stored memories that consist of the previous statements in the story. 
The original proposed model performs {\em 2 hops} of inference: finding the first supporting fact
with the maximum match score with the question, and then the second supporting fact with the maximum match score with both the question and the first fact that was found. 
The matching function consists of mapping the bag-of-words for the question and facts into an embedding space by summing word embeddings. The word embeddings are learnt using strong supervision 
to optimize the QA task.
After finding supporting facts, a final ranking is performed to rank possible responses (answer words) given those facts.
We also consider some extensions to this model:
\begin{itemize}
\item {\bf Adaptive memories} performing a variable number of hops rather than 2, the model is trained to predict a hop or the special ``STOP'' class. A similar procedure can be applied to output multiple tokens as well.
\item {\bf $N$-grams} We tried using a bag of 3-grams rather than a bag-of-words to represent the text. In both cases the first step of the MemNN is to convert these into vectorial embeddings.
\item {\bf Nonlinearity} We apply a classical 2-layer neural network with tanh nonlinearity in the matching
function.
\end{itemize}
More details of these variants is given in Sec \ref{sec:memnns} of the appendix.


Finally, we  built a classical cascade NLP system baseline using a structured support vector 
machine (SVM), which 
incorporates coreference resolution and semantic role labeling (SRL) preprocessing steps,
which are themselves trained on large amounts of costly labeled data.
The Stanford coreference system \citep{raghunathan2010multi} 
and the SENNA semantic role labeling  (SRL) system \citep{collweston} are used to build
 features for the input to the SVM, trained with strong supervision to 
find the supporting facts, e.g. features based on words, word pairs, and 
the SRL verb and verb-argument pairs.
After finding the supporting facts, we build a similar structured SVM for the response stage, with features
tuned for that goal as well. 
More details are  in Sec. \ref{sec:exp-nlp} of the appendix.

Learning rates and other hyperparameters for all methods 
are chosen using the training set. 
The summary of our experimental results on the tasks is given in Table \ref{table:results}.
We give results for each of the 20 tasks separately, as well as  mean performance and number of failed tasks
 in the final two rows.

\paragraph{Results}
Standard MemNNs generally outperform the \N-gram and LSTM baselines, which is consistent with the results
in \mbox{\citet{memnns}}. However they still ``fail'' at a number of tasks; that is, 
test accuracy is less than 95\%.
Some of these failures are expected due to insufficient modeling power as
 described in more detail in Sec. \ref{sec:memnn-shortcomings}, e.g. $k=2$ facts, single word answers and bag-of-words do not succeed on tasks \secthreefacts, \sectwoargrel,
 \secthreeargrel, \seccount, \secsets~and~\secszreas. However, there were also failures on tasks we did not at first expect, for example yes/no questions
(\secyesno) and indefinite knowledge (\secindefreason). Given hindsight, we realize that the
linear scoring function of  standard MemNNs cannot model the match between query, supporting fact and a yes/no answer as this requires three-way interactions.

Columns 5-9 of Table \ref{table:results} give the results for our MemNN extensions: 
adaptive memories (AM), \N-grams (NG)  and nonlinearities (NL), plus combinations thereof.
The adaptive approach gives a straight-forward improvement in tasks \secthreefacts~
 and \secinduction~
because they both require more than two supporting facts, and also
gives (small) improvements in \secsets~ and \secpathfinder~because they require multi-word outputs
(but still remain difficult). We hence use the AM model in combination with all our other extensions in the subsequent experiments.

MemNNs with \N-gram modeling yield clear improvements when word order matters, e.g. tasks \sectwoargrel
 ~and~\secdeduction. However, \N-grams 
 do not seem to be a substitute for nonlinearities in the embedding
function as the NL model outperforms \N-grams on average, especially in the yes/no (\secyesno) and 
indefinite tasks (\secindefreason), as explained before. On the other hand, the NL method cannot model word order and so fails e.g., on task \sectwoargrel. The obvious step is thus to combine these complimentary approaches: indeed AM+NG+NL (column 9) gives improved results over both, with a total of 9 tasks that have been upgraded from failure to success compared to the original MemNN model. 

The structured SVM, despite having access to external resources, does not perform better, 
still failing at 9 tasks.
It does perform better than vanilla MemNNs (without extensions)
on tasks \secyesno,~\secnegation~and \secindefreason~where
the hand-built feature conjunctions capture the necessary nonlinearities.
However, compared to MemNN (AM+NG+NL)
it seems to do significantly worse on tasks requiring three (and sometimes, two)
 supporting facts
(e.g. tasks \secthreefacts, \secinduction~and \sectwofacts)
presumably as ranking over so many possibilities introduces more mistakes.
However, its non-greedy search does seem to help on other tasks,
such as path finding (task \secpathfinder)
 where search is very important.
Since it relies on external resources specifically designed for
English, it is unsure that it would perform as well on other
languages, like Hindi, where such external resources might be of worse quality.

The final two columns (10-11) give further analysis of the AM+NG+NL MemNN method. 
The second to last column (10) shows the minimum number of training examples required to achieve $\geq$ 95\% accuracy, or {\em FAIL} if this is not achieved with 1000 examples.
This is important as it is not only desirable to perform well on a task, but also using the fewest number of examples
(to generalize well, quickly).
Most succeeding tasks require 100-500 examples. Task \secsets~ requires 5000 examples and \seccount~
 requires 10000, hence they are labeled as {\em FAIL}.
The latter task can presumably be solved by adding all the times an object is picked up, and subtracting the times it is dropped, which seems possible for an MemNN, but it does not do perfectly. 
Two tasks, positional reasoning \secposreas~and path finding \secpathfinder~ cannot be solved even with 10000 examples, it seems those (and indeed more advanced forms of induction and deduction, which we plan to build) require a general search algorithm to be built into the inference procedure, which MemNN (and the other approaches tried) are lacking.

The last column shows the performance of AM+NG+NL MemNNs
 when training on {\em all} the tasks jointly, rather than just on a single one.
The performance is generally encouragingly similar, showing such a model can learn many aspects of text
understanding and reasoning simultaneously. 
The main issues are that these models still fail on several
of the tasks, and use a far stronger form of supervision (using supporting facts) than is typically realistic.

%% file: Conclusion.tex
\paragraph{A prerequisite set} We developed a set of tasks that we believe are a prerequisite to full language understanding and reasoning.
While any learner that can solve these tasks is not necessarily close to
full reasoning, if a learner fails on any of our tasks then
there are likely real-world tasks
that it will fail on too
(i.e., real-world tasks that require the same kind of reasoning).
%
%
Even if the situations and the language of the tasks are artificial,
we 
believe that the mechanisms required to {\em learn} how to solve them are part of the key towards
text understanding and reasoning.

\vspace{-2mm}
\paragraph{A flexible framework} This set of tasks is not a definitive
set. The purpose of a simulation-based approach is to provide
flexibility and 
 control of the tasks' construction. 
We grounded the tasks 
into language because 
it is then easier to understand the usefulness of the tasks and to interpret their results.
However,  our primary
goal is to find models able to learn to detect and
combine patterns in symbolic sequences.
One might even want to 
  decrease the intrinsic
difficulty by removing any lexical variability and ambiguity and
reason only over bare symbols, stripped down from their linguistic meaning.
One could also decorrelate the long-term memory from the reasoning capabilities
of systems by, for instance,
arranging the supporting facts closer to the questions.
%
In the opposing view, one could instead want to transform the tasks into
more realistic stories using annotators or more complex grammars.
The set of 20 tasks presented here is a subset of what can be achieved
with a simulation. We chose them because they offer a variety of
skills that we would like a text reasoning model to have, but we hope
researchers from the community will develop more tasks of varying complexity
in order to develop and analyze models that try to solve them.
Transfer learning across tasks is also a very important goal, beyond the scope of this paper.
We have thus made the simulator and code for the tasks publicly available for those purposes.

\vspace{-2mm}
\paragraph{Testing learning methods} Our tasks are designed as a
test-bed for learning methods: we provide training and test sets
because we intend to evaluate the capability of models to discover how to
reason from patterns hidden within them. 
It could be tempting to hand-code solutions for them or to use 
existing large-scale QA systems like 
Cyc \citep{curtis2005effective}.
They might succeed at solving them,
even if our structured SVM results (a cascaded NLP system with hand-built features)
  show that this is not 
straightforward; however this is not the tasks' purpose since those 
approaches would not be learning to solve them.
Our experiments show that  some existing machine learning methods
 are successful on some of the tasks, in particular Memory Networks, for which
we introduced some useful extensions (in Sec. \ref{sec:memnns}).
However, those models still fail on several
of the tasks, and use a far stronger form of supervision (using supporting facts)
than is typically realistic.

These datasets are not yet solved.
Future research should aim to minimize the amount of required supervision,
as well as the number of training examples needed to solve a new task, 
to move closer to the task transfer capabilities of humans.
That is, in the weakly supervised case with only 1000 training  examples or less 
there is no known general (i.e. non-hand engineered) method that
solves the tasks.
Further, importantly, our hope is that a feedback loop of developing more challenging tasks,
and then algorithms that can solve them, leads us to fruitful research directions.

Note that these tasks are not a substitute for real data, but should complement
them, especially when developing and analysing algorithms. There are many complementary
real-world datasets, see for example  \cite{nips15_hermann,bordes2015large,hill2015goldilocks}.
That is, even if a method works well on our 20 tasks, it should be shown to be useful on
 real data as well.


\paragraph{Impact}
Since being online, the bAbI tasks have already directly influenced the development of several
promising new algorithms, including weakly supervised end-to-end Memory Networks (MemN2N) 
of \citet{sukhbaatar2015end},
Dynamic Memory Networks of \cite{kumar2015ask},
and the Neural Reasoner \citep{peng2015towards}.
MemN2N has since been shown to perform well on some real-world tasks
\citep{hill2015goldilocks}.

%% file: Models.tex
Memory Networks \cite{memnns} are a promising class of models, shown to perform well at QA,
that we can apply to our tasks.
%
%
They consist of a memory $\m$ (an array of objects
indexed by $\m_i$) and
 four potentially learnable components 
 $I$, $G$, $O$ and $R$
that are executed given an input: 

\vspace{-1mm}
\begin{itemize}
\vspace{-1mm}
\item[I:] (input feature map) -- convert input sentence $x$ to an internal feature representation $I(x)$.
\vspace{-1mm}
\item[G:] (generalization) -- update the current memory state $\m$ given the new input: ~ $\m_i  = G(\m_i, I(x), \m), ~ \forall i$.
\vspace{-1mm}
\item[O:] (output feature map) -- compute output $o$ given the new input and the memory: $o  = O(I(x), \m)$.   
\vspace{-1mm}
\item[R:] (response) -- finally, decode output features $o$ to give the final textual response to the user:  $r = R(o)$.
\end{itemize}

\vspace{-1mm}
Potentially, component $I$ can make use of standard
pre-processing, e.g., parsing and entity resolution,
 but the simplest form is to do no processing at all.
The simplest form of $G$ is store the new incoming example in an empty memory slot, and leave the rest of the memory untouched.
Thus, in \cite{memnns} the actual implementation used is exactly this simple form, where the bulk of the work is in the $O$ and $R$ components. The former is responsible for reading 
from memory and performing inference,
e.g., calculating what are the relevant memories to answer a question,
and the latter for producing the actual wording of the answer given $O$.

The $O$ module produces output features by finding $k$ supporting
memories given $x$. They use $k=2$. 
For $k=1$ the highest scoring supporting memory is retrieved with:
\begin{equation}
      o_1 = O_1(\inp, \m) =  \argmax\limits_{i=1,\dots,N} ~s_O(\inp, \m_i)
\label{eq:o1}
\end{equation}
where $s_{O}$ is a function that scores the match between the pair of sentences
$x$ and $\m_i$. 
For the case $k = 2$ they then find a second supporting memory given the first found in the previous iteration:
\begin{equation}
      o_2 = O_2(q, \m) =   \argmax\limits_{i=1,\dots,N}  ~s_O([\inp, \m_{o_1}], \m_i)
\label{eq:o2}
\end{equation}
where the candidate supporting memory $\m_i$ is now scored with respect to both the original
input and the first supporting memory, where square brackets
denote a list.
The final output $o$ is $[x, \m_{o_1}, \m_{o_2}]$, which is input to the module $R$.

Finally, $R$ needs to produce a textual response $r$.
While the authors also consider Recurrent Neural Networks (RNNs),
their standard setup 
limits responses to be a single word (out of all the words
seen by the model) by ranking them:
\begin{equation} \label{eq:R-word-score}
    r  =  R(q, w) = {\mbox{argmax}}_{w \in W}  ~s_R([\inp, \m_{o_1}, \m_{o_2}], w)
\end{equation}
where $W$ is the set of all words in the dictionary, and $s_R$ is a
function that  scores the match.

The scoring functions 
$s_O$ and $s_R$ have the same form, that of an embedding model:
\begin{equation}
  s(x,y) = \Phi_x(x)^\top U^\top U \Phi_y(y).
\end{equation}
where $U$ is a $n \times D$ matrix where $D$ is the number of features
and $n$ is the embedding dimension.  The role of
$\Phi_x$ and $\Phi_y$ is to map the original text to the
$D$-dimensional feature space.  
They choose  a bag of words representation, and  $D = 3|W|$ for $s_O$,
i.e., every word in the dictionary has three different representations:
one for $\Phi_y(.)$ and two for $\Phi_x(.)$ depending on whether the
words of the input arguments are from the actual input $x$ or from
the supporting memories 
so that they can be modeled
differently.


They consider various extensions of their model,
in particular modeling write time and modeling unseen words. Here we only discuss the former which we also use.
In order for the model to work on QA tasks over stories it needs to know which
order the sentences were uttered which is not available in the model directly.
They thus add extra write time extra features to $S_{O}$  
which take on the value 0 or 1 indicating which sentence is older than another being compared,
and compare triples of pairs of sentences and the question itself. 
Training is carried out by stochastic gradient descent using supervision from both the question answer pairs and the supporting memories  (to select $o_1$ and $o_2$).
See \cite{memnns} for more details.

\subsection{Shortcomings of the Existing MemNNs} \label{sec:memnn-shortcomings}
The Memory Networks models defined in \citep{memnns} are one possible technique to try on our
tasks, however there are several tasks which they are likely to fail on:
\begin{itemize}
\item \vspace{-3mm} They model sentences with a bag of words so are likely to fail on tasks such as
the 2-argument (task \sectwoargrel) and 3-argument (task \secthreeargrel) relation problems.
\item They perform only two max operations ($k=2$) so they cannot handle questions involving more than two supporting facts such as tasks \secthreefacts ~and \seccount.
\item Unless a RNN is employed in the R module, they are unable to provide multiple answers in the standard setting using eq. (\ref{eq:R-word-score}). This is required for the 
 list (\secsets)  and  path finding (\secpathfinder) tasks.
\end{itemize}
\vspace{-3mm}
We therefore propose improvements to their model in the following section.

\subsection{Improving Memory Networks} \label{sec:memext}

\subsubsection{Adaptive Memories (and Responses)} \label{sec:adap}
We consider a variable number of supporting facts that is automatically adapted dependent on the question being asked.
To do this we consider scoring a special fact $m_\emptyset$. Computation of supporting memories then becomes:

\begin{small}
\begin{algorithmic}
\STATE $i = 1$
\STATE $o_i = O(x, \m)$
\WHILE{$o_i \neq  m_\emptyset$}
\STATE $i \leftarrow i + 1$
\STATE $o_i = O([x, m_{o_1}, \dots, m_{o_{i - 1}}], \m)$
\ENDWHILE 
\end{algorithmic}
\end{small}
That is, we keep predicting supporting facts $i$, conditioning at each step on the previously found facts, until 
$m_\emptyset$ is predicted at which point we stop. $m_\emptyset$ has its own unique embedding vector, which is also learned. In practice we still impose a hard maximum number of loops in our experiments to avoid fail cases where the computation never stops (in our experiments we use a limit of 10).

\vspace{-2mm}
\paragraph{Multiple Answers}
We use a similar trick for the response module as well in order to output multiple words.
That is, we add a special word $w_\emptyset$ to the dictionary and predict word $w_i$ on each iteration $i$
conditional on the previous words, i.e., 
$w_i = R([x, m_{o_1}, \dots, m_{|o|}, w_i, \dots, w_{i-1}], w)$, until
we predict $w_\emptyset$.

\subsubsection{Nonlinear Sentence Modeling} \label{sec:nonlin}
 There are several ways of modeling sentences that go beyond a bag-of-words, and we explore three variants here.
The simplest is a bag-of-{\bf \N-grams}, we consider $N = 1, 2$ and $3$ in the bag. The main disadvantage of such a method is
that the dictionary grows rapidly with $N$. We therefore consider an alternative neural network approach, which we call
a {\bf multilinear} map. Each word in a sentence is binned into one of $P_{sz}$ positions with $p(i, l) = \ceil*{(iP_{sz})/l)}$ where $i$ is the position of the word in a sentence of length $l$, and for each position we employ a $n \times n$ matrix $P_{p(i, l)}$. We then model the matching score with:
\begin{equation}
\small  s(q,d) = E(q) \cdot E(d);~~~  E(x) =    tanh(\sum_{i=1,\dots,l} P_{p(i, l)} \Phi_x(x_i)^\top U)
\end{equation}
whereby we apply a linear map for each word dependent on its position, followed by a $tanh$ nonlinearity on the sum of 
mappings. Note that this is related to the model of \citep{yufactor} who consider tags rather than positions.
While the results of this method are not shown in the main paper due to space
restrictions, it performs similarly well to $N$-grams
 to and may be useful in real-world cases where \N-grams
cause the dictionary to be too large. Comparing to
Table \ref{table:results}
 MemNN with adaptive memories (AM) + multilinear 
 obtains a mean performance of 93, the same as MemNNs with AM+NG+NL (i.e., using 
 N-grams instead). 

Finally, to assess the performance of nonlinear maps that do not model word position at all we also consider the following {\bf nonlinear} embedding:

\begin{equation}
 \small   E(x) =    tanh(W tanh(\Phi_x(x)^\top U)).
\end{equation}
where $W$ is a $n \times n$ matrix.
This is similar to a classical two-layer neural network, but applied to both sides $q$ and $d$ of $s(q,d)$.
We also consider the straight-forward combination of bag-of-\N-grams followed by this nonlinearity.

%% file: NLPBaseline.tex
We also built a classical cascade NLP system  baseline using a structured SVM, which 
incorporates  coreference resolution and semantic role labeling preprocessing steps,
which are themselves trained on large amounts of costly labeled data. 
We first run the Stanford coreference system \citep{raghunathan2010multi} on the stories
and each mention is then replaced with the first mention of its entity class.
Second, the SENNA semantic role labeling system (SRL) \citep{collweston} is run, and we collect the set
of arguments for each verb. 
We then define a ranking task for finding the supporting facts (trained using strong supervision):
\[
  o_1,o_2,o_3 = \argmax_{o \in {\cal O}}{S_O(x, f_{o_1}, f_{o_2}, f_{o_3}; \Theta)}  
\]
where given the question $x$ 
we find at most three supporting facts with indices $o_i$ from the set of facts $f$ in the story
(we also consider selecting an ``empty fact''  for the case of less than three),
and $S_O$ is a linear scoring function with parameters $\Theta$.
Computing the argmax requires doing exhaustive search, unlike e.g. the MemNN method which is greedy.
For scalability,
we thus prune the set of possible matches by requiring that facts share one common non-determiner word
with each other match or with $x$.
$S_O$ is constructed as a set of indicator features. For simplicity
each of the features only looks at pairs of sentences, i.e.
$S_O(x, f_{o_1}, f_{o_2}, f_{o_3}; \Theta) = $  $\Theta * (g(x,f_{o_1}), g(x, f_{o_2}),  g(x, f_{o_3}), g(f_{o_1},f_{o_2}), g(f_{o_2},f_{o_3}),$ $g(f_{o_1},f_{o_3}))$.
The feature function $g$ is made up of the following feature types, shown here for $g(f_{o_1},f_{o_2})$:
(1) Word pairs: One indicator variable for each pair of words in $f_{o_1}$ and $f_{o_2}$.
(2) Pair distance: Indicator for the distance between the sentence, i.e. $o_1 - o_2$.
(3) Pair order: Indicator for the order of the sentence, i.e. $o_1 > o_2$.
(4) SRL Verb Pair: Indicator variables for each pair of SRL verbs in $f_{o_1}$ and $f_{o_2}$.
(5) SRL Verb-Arg Pair: Indicator variables for each pair of SRL arguments in $f_{o_1}$, $f_{o_2}$ and their 
corresponding verbs.
After finding the supporting facts, we build a similar structured SVM for the response stage, also with features
tuned for that goal:  Words -- indicator for each word in $x$,
 Word Pairs -- indicator for each pair of words in $x$ and supporting facts,
and similar SRL Verb and SRL Verb-Arg Pair features as before.

Results are given in Table \ref{table:results}.
The structured SVM, despite having access to external resources, does not perform better than MemNNs overall, still failing at 9 tasks.
It does perform well on tasks \secyesno,
\secnegation ~and \secindefreason where
the hand-built feature conjunctions capture the necessary nonlinearities that the original MemNNs 
do not.
However, it seems to do significantly worse on tasks requiring three (and sometimes, two)
 supporting facts
(e.g. tasks \secthreefacts, \secinduction ~and \sectwofacts)
presumably as ranking over so many possibilities introduces more mistakes.
However, its non-greedy search does seem to help on other tasks,
such as path finding (task \secpathfinder)  where search is very important.